\documentclass[11pt,a4paper]{article}
\usepackage[hyperref]{emnlp2018}
\usepackage{times}
\usepackage{latexsym}
\usepackage{xspace}
\usepackage[colorinlistoftodos]{todonotes} 
\usepackage{multirow}
\usepackage{url}
\usepackage{fixltx2e}

\aclfinalcopy 

\newcommand{\sref}[1]{Section$\,$\ref{#1}}
\newcommand{\cref}[1]{Chapter$\,$\ref{#1}}
\newcommand{\tref}[1]{Table$\,$\ref{#1}}
\newcommand{\fref}[1]{Figure$\,$\ref{#1}}

\newcommand{\fone}{F$_1$\xspace}

\newcommand{\nombank}{\texttt{NomBank}\xspace}
\newcommand{\pcedt}{\texttt{PCEDT}\xspace}

\title{Transfer and Multi-Task Learning for\\[0.3ex]
       Noun--Noun Compound Interpretation} 

\author{Murhaf Fares
  \\\And
  Stephan Oepen \\ \\
  Department of Informatics \\
  University of Oslo \\
  {\tt {murhaff|oe|erikve}@ifi.uio.no}
   \\\\\And
  Erik Velldal \\}

\date{}

\begin{document}
\maketitle
\begin{abstract}

In this paper, we empirically evaluate the
utility of transfer and multi-task learning on a challenging semantic
classification task: semantic interpretation of noun--noun compounds.
Through a comprehensive series of experiments and in-depth error analysis,
we show that transfer learning via parameter
initialization and multi-task learning via parameter sharing can help a
neural
classification model generalize over a highly skewed distribution of relations.
Further, we demonstrate how dual annotation with two distinct sets of relations
over the same set of compounds can be exploited to improve the overall accuracy
of a neural classifier and its \fone scores on the less frequent, but more
difficult relations.

\end{abstract}

\section{Introduction}
\label{sec:intro}

Noun--noun compound interpretation is the task of assigning semantic
relations to pairs of nouns (or more generally, pairs of noun phrases in the
case of multi-word compounds). 
For example, given the nominal compound \emph{street protest}, the task of
compound interpretation is to predict the semantic relation holding between 
\emph{street} and \emph{protest} (a locative relation in this example).
Given the frequency of noun--noun compounds in natural language -- e.g.\ 3\% of
the tokens in the British National Corpus \cite{Burnard:95} are part of
noun--noun compounds \cite{Sea:08} -- and its relevance to other natural language
processing (NLP) tasks such as question answering and information retrieval
 \cite{Nakov:08}, noun--noun compound interpretation has been the focus of much work,
in theoretical linguistics 
\cite{Li:72,Downing:77,Levi:78,Finnin:80,Ryder:94},
psycholinguistics \cite{Gag:Sho:97,Mar:Gag:Spa:17},
and computational linguistics
\cite{Lauer:95,Nakov:07,Sea:Cop:09,Gir:Nak:Nas:09,Kim:Bal:13,Dim:Hin:15}. 
In computational linguistics, noun--noun compound interpretation is, by and
large, approached as an automatic classification problem. 
Hence several machine learning (ML) algorithms and
models have been used to learn the semantics of nominal compounds, including
Maximum Entropy \cite{Tra:Hov:10}, Support Vector Machines \cite{Sea:Cop:13} and
Neural Networks \cite{Dim:Hin:15,Ver:Wat:18}. 
These models use information from lexical semantics such as WordNet-based
features and distributional semantics such as word embeddings.
Nonetheless, noun--noun compound interpretation remains one of the more
difficult NLP problems because: 1) noun--noun compounding, as a linguistic
construction, is very productive and 2) the semantics of noun--noun compounds is
not easily derivable from the compounds' constituents \cite{Ver:Wat:18}.
Our work, in part, contributes to advancing NLP research on noun--noun
compound interpretation through the use of transfer and multi-task learning.

The interest in using transfer learning (TL) and multi-task learning (MTL) in
NLP has surged over the past few years, showing `mixed'
results depending on the so-called main and auxiliary tasks involved, model
architectures and datasets, among other things
\cite{Col:Wes:08,Mou:Men:Yan:16,Sog:Gold:16,Alo:Pla:17,Bin:Sog:17}.
These `mixed' results, coupled with the fact that neither TL nor MTL has been
applied to noun--noun compounds interpretation before, motivate our
extensive empirical study on the use of TL and MTL for compound
interpretation, not only to supplement existing research on the utility of
TL and MTL for semantic NLP tasks in general, but also to determine their
benefits for compound interpretation in particular.

One of the primary motivations for using multi-task learning is to improve
generalization by ``leveraging the domain-specific information contained in the
training signals of \emph{related} tasks'' \newcite{Caruana:97}.
In this work, we show that TL and MTL can indeed be used as a kind of 
\emph{regularizer} to learn to predict infrequent relations given a highly
skewed distribution of relations from the noun--noun compound dataset of 
\newcite{Fares:2016} which is especially well suited for TL and MTL
experimentation as detailed in \sref{sec:task}.

Our contributions can be summarized as:
\begin{enumerate}
  \item Through careful result analysis, we find that TL and MTL (mainly on
  the embedding layer) do improve the overall accuracy and the \fone scores of
  the less frequent relations in a highly skewed dataset, in comparison to a
  strong single-task learning baseline.
  \item Even though our work focuses on TL and MTL, to the
  best of our knowledge, we are the first to report experimental results on the
  comparatively recent dataset of \newcite{Fares:2016}.
\end{enumerate}

\section{Related Work}
\label{sec:related}

\paragraph{Noun--Noun Compound Interpretation} 
Existing approaches to noun--noun compound interpretation vary depending on
the taxonomy of compound relations as well as the machine learning models and
features used to learn those relations.
For example, \newcite{Sea:07} defines a coarse-grained set of relations (viz.\
six relations based on theoretical work by \newcite{Levi:78}), whereas 
\newcite{Tra:Hov:10} assume a considerably more fine-grained taxonomy of
43 relations. 
Others question the very assumption that noun--noun compounds are interpretable
using a finite, predefined set of relations \cite{Downing:77,Finnin:80} and
propose alternative paraphrasing-based approaches \cite{Nakov:07,Shw:Dag:18}.
We here focus on the approaches that cast the interpretation problem as a 
classification task over a finite predefined set of relations.
A wide variety of machine learning models have been already applied to learn
this task, including nearest neighbor classifiers using semantic similarity based
on lexical resources \cite{Kim:Bal:05}, kernel-based methods like SVMs using
lexical and relational features \cite{Sea:Cop:09}, Maximum Entropy models with a
relatively large selection of lexical and surface form features such as synonyms
and affixes \cite{Tra:Hov:10} and, most recently, neural networks
either solely relying on word embeddings to represent noun--noun compounds 
\cite{Dim:Hin:15} or word embeddings and so-called path embeddings (which encode
information about lemmas and part-of-speech tags, inter alia) in a combined
paraphrasing and classification approach \cite{Ver:Wat:18}.
Of the aforementioned studies, \newcite{Tra:Hov:10,Dim:Hin:15,Ver:Wat:18}
have all used the same dataset by \newcite{Tra:Hov:10}. 
To the best of our knowledge, TL and MTL have never been applied to compound
interpretation before, and in the following we therefore review some of the 
previous work on TL and MTL on other NLP tasks.

\paragraph{Transfer and Multi-Task Learning} A number of recent studies have presented
comprehensive experiments on the use of TL and MTL for a variety of NLP
tasks including named entity recognition and semantic labeling 
\cite{Alo:Pla:17}, sentence-level sentiment classification \cite{Mou:Men:Yan:16},
super-tagging and chunking \cite{Bin:Sog:17} and semantic dependency parsing 
\cite{Pen:Tho:Smi:17}.
The common thread among the findings of these studies is that the benefits of
TL and MTL largely depend on the properties of the tasks at hand, such as the
skewedness of the data distribution \cite{Alo:Pla:17}, the semantic similarity
between the source and target tasks \cite{Mou:Men:Yan:16}, the learning
pattern of the auxiliary and main tasks where ``target tasks that quickly
plateau'' benefit most from ``non-plateauing auxiliary tasks'' 
\cite{Bin:Sog:17} and the ``structural similarity'' between the tasks 
\cite{Pen:Tho:Smi:17}.
In addition to the difference in the NLP tasks they experiment with, the
aforementioned studies assume slightly different definitions of TL and MTL 
(cf.\ \sref{sec:tl-vs-mtl}).
Our work is similar in spirit to that of \newcite{Pen:Tho:Smi:17} in the sense
that we use TL and MTL to learn different `formalisms' (semantic annotations of
noun--noun compounds in our case) on the \emph{same} dataset.
However, our experimental setup is more similar to the work by 
\newcite{Mou:Men:Yan:16} in that we experiment with parameter
initialization on all the layers of the neural model and simultaneously train
one MTL model on two sets of relations (cf.\ \sref{sec:models}). 

\section{Task Definition and Dataset}
\label{sec:task}

Given a set of labeled pairs of nouns, each a noun--noun compound, the
task is simply to learn to classify the semantic relations holding between each
pair of compound constituents.  
The difficulty of this task, obviously, depends on the label set used and its
distribution, among other things.
For all the experiments presented in this paper, we adapt the noun--noun
compounds dataset created by \newcite{Fares:2016} which consists of compounds
annotated with two different taxonomies of relations; in other words, for each
noun--noun compound there are two distinct relations, drawing on different
linguistic schools.
The dataset was derived from existing linguistic resources, such as
\nombank \cite{Mey:Ree:Mac:04} and the Prague Czech-English Dependency Treebank
2.0 \cite[\pcedt]{Haj:Haj:Pan:12}.
Our motivation for using this dataset is twofold: first, dual annotation with
relations over the same underlying set of compounds maximally enables TL and MTL
perspectives; second, alignment of two distinct annotation frameworks over the
same data facilitates contrastive analysis and comparison across frameworks.

More specifically, we use a subset of the dataset created by
\newcite{Fares:2016}, by focusing on type-based instances of so-called 
two-word compounds.\footnote{Two-word compounds consist of two
whitespace-separated constituents. 
A single constituent, however, can be a `simple' noun (e.g.\ \emph{system}) or
a hyphenated noun (e.g.\ \emph{land-ownership}) leading to compounds like
\emph{land-ownership system}. The representation of compounds with hyphenated
constituents is explained in \sref{sec:stl}}
The original dataset by \newcite{Fares:2016} also includes multi-word compounds 
(i.e.\ compounds consisting of more than two nouns) and more than just one
instance per compound type.
Furthermore, we define a three-way split of the dataset; 
\tref{tab:dataset-stats} presents the number of compound types per split and the
vocabulary size of each split (i.e.\ the number of unique words in each split);
the latter is also broken down in terms of words occurring in the
right-most position (right constituents) and the left-most position 
(left constituents).\footnote{We use the terms left and
right constituents, instead of modifier and head nouns, because the dataset
does not make explicit the notion of `headedness'.}
Overall, the two label sets consists of 35 so-called \pcedt functors and 18
\nombank argument and adjunct relations.
As detailed in \sref{ssec:analysis-dataset},
these label sets are far from being uniformly distributed.

\begin{table}[t!]
\small
\begin{center}
\begin{tabular}{@{}l | c | c | c@{}}
    & Train & Dev & Test \\ \hline
Compounds  & 6932 & 920 & 1759 \\
Vocab size  & 4102 & 1163 & 1772 \\
Right constituents & 2304 & 624 & 969 \\
Left constituents & 2405 & 618 & 985 \\
\hline
\end{tabular}
\end{center}
\caption{\label{tab:dataset-stats} Characteristics of the noun--noun
compound dataset used in our experiments. The numbers in this table
correspond to a (sub)set of the dataset by \newcite{Fares:2016}, see 
\sref{sec:task}.}
\end{table}

Abstractly, many relations in \pcedt and \nombank describe
similar semantic concepts, since they annotate the semantics of the same text.
For example, \newcite{Fares:2016} reports that 
the temporal and locative relations in \nombank 
(\texttt{ARGM-TMP} and \texttt{ARGM-LOC}, respectively) and their
counterparts in \pcedt (\texttt{TWHEN} and \texttt{LOC}) exhibit a relatively
consistent behavior across frameworks as they annotate many of the same
compounds.
However, \newcite{Fares:2016} also points out that some abstractly similar
relations do not align well in practice; for example, the functor \texttt{AIM}
in \pcedt and the modifier argument \texttt{ARGM-PNC} in \nombank express a
somewhat similar semantic concept (purpose) but the overlap between the
sets of compounds they annotate in practice is rather small.
Nonetheless, it is plausible to assume that the semantic similarity in the label
sets---whenever it exists---can be exploited in the form of transfer and
multi-task learning, not least because the overall distribution of the relations
in the two frameworks is different.

\section{Transfer vs.\ Multi-Task Learning}
\label{sec:tl-vs-mtl}

In this section, we use the notations and definitions by \newcite{Pan:Yan:10} to
define our setup for transfer and multi-task learning.

Our classification task $\mathcal{T}$ can be defined in
terms of all training pairs $(X, Y)$ and a probability distribution $P(X)$, where
$X = {x_i,\dots,x_N} \in \mathcal{X}$ and $Y = {y_i,\dots,y_N} \in \mathcal{Y}$;
$\mathcal{X}$ is the input feature space, $\mathcal{Y}$ is the set of all
labels and $N$ is the size of the training data.
A task's domain $\mathcal{D}$ is defined by $\left\{\mathcal{X},P(X)\right\}$.
Our goal is to learn a function $f(X)$ that predicts $Y$ based on the input
features $X$.
Assuming two ML tasks, $\mathcal{T}_a$ and $\mathcal{T}_b$, 
we would train two models (i.e.\ learn two separate functions $f_a$ and
$f_b$) to predict $Y_a$ and $Y_b$ in a single-task learning setup.
However, if $\mathcal{T}_a$ and $\mathcal{T}_b$ are related somehow, either
explicitly or implicitly, TL and MTL can improve the generalization of either
task or both \cite{Caruana:97,Pan:Yan:10,Mou:Men:Yan:16}.
Two tasks are considered related when their domains, 
$\mathcal{D}_a$ and $\mathcal{D}_b$, are similar but their label sets are
different $\mathcal{Y}_a \neq \mathcal{Y}_b$ or when their domains are different
but their label sets are identical, i.e.\ $\mathcal{Y}_a = \mathcal{Y}_b$ 
\cite{Pan:Yan:10}.\footnote{When the label sets are identical,
TL practically becomes a technique for \emph{domain adaptation}. 
Though these two terms have also been used interchangeably 
\cite{Chu:Lee:Hun:18}.}
As such, noun--noun compound interpretation over the dataset of
\newcite{Fares:2016} is a well suited candidate for TL and MTL, because the
training examples are identical, i.e.\ 
$X_{\scriptscriptstyle PCEDT} = X_{\scriptscriptstyle NomBank}$, but the label
sets are different 
$\mathcal{Y}_{\scriptscriptstyle PCEDT} \neq \mathcal{Y}_{\scriptscriptstyle NomBank}$.

For the sake of clarity, we distinguish between transfer learning and
multi-task learning in this paper, even though these two terms are at times
used somewhat interchangeably in the literature. For example, what we call TL and MTL in this
paper are both referred to as transfer learning by \newcite{Mou:Men:Yan:16}.
We define TL as using the parameters (i.e.\ weights in neural networks) of one
model trained on $\mathcal{T}_a$ to initialize another model for 
$\mathcal{T}_b$. \newcite{Mou:Men:Yan:16} refer to this method
as ``parameter initialization''.\footnote{Using \emph{pretrained} word embeddings
as input representation is in a sense a form of unsupervised transfer
learning, but in this work we focus on transfer learning based on supervised
learning.} 
MTL, on the other hand, here refers to training (parts of) the same model to learn
$\mathcal{T}_a$ and $\mathcal{T}_b$, i.e.\ learning one set of parameters for
both tasks. 
The idea is to train a single model simultaneously
on the two tasks where one task is considered to introduce inductive bias which
would help the model generalize over the main task. 
Note, however, that this does not necessarily mean that we eventually want to
use a single model to predict both label sets in practice (cf.\ \sref{ssec:mtl}). 

\section{Neural Classification Models}
\label{sec:models}

Here we present the neural classification models used in our experiments.
To isolate the effect of TL and MTL, we first present a single-task learning
model, which serves as our baseline model, and then we use the same model to
apply TL and MTL.

\subsection{Single-Task Learning Model}
\label{sec:stl}

In our single-task learning (STL) setup, we train and fine-tune a feed-forward
neural
network based on the neural classifier proposed by  
\newcite{Dim:Hin:15}, which consists of: 1) input
layer, 2) embedding layer, 3) hidden layer, and 4) output layer.
The input layer is simply two integers specifying the indices of a 
compound's constituents in the embedding layer where the word embedding vectors
are stored; the selected word embedding vectors are then fed to a fully
connected hidden layer whose size is the same as the number of dimensions of
the word embedding vectors.
Finally, a \emph{softmax} function is applied on the output layer and the most
likely relation is selected.

The compound's constituents are represented using a $300$-dimensional word
embedding model trained on an English Wikipedia dump 
(dated February 2017) and English Gigaword Fifth Edition \cite{Par:Gra:Kon:11}
using GloVe \cite{Pen:Soc:Man:14}. 
The embedding model was trained by \newcite{Far:Kut:Oep:17} who provide more
details on the hyperparameters used to train the embedding 
model.\footnote{\href{http://vectors.nlpl.eu/}{vectors.nlpl.eu/repository}}
When looking up a word in the embedding model, if the word is not found
we check if the word is uppercased and look up the same word in lowercase.
If a word is hyphenated and is not found in the embedding vocabulary, we split it on the
hyphen and average the vectors of its parts (if they exist in the vocabulary).
If after these steps the word is still not found, we use a designated vector
for unknown words.

\paragraph{Architecture and Hyperparameters}
Our choice of hyperparameters is motivated by several rounds of experimentation
on the single-task learning model as well as the choices made by 
\newcite{Dim:Hin:15}. 

The weights of the embedding layer (i.e.\ the word embeddings) are
updated during training in all the models.
The optimization function we use in all the models is Adaptive Moment
Estimation, known as \emph{Adam} \cite{Kin:Ba:15} with $\eta=0.001$ (the
default learning rate). 
The loss function is negative-log likelihood (aka categorical cross-entropy).
We use a \emph{Sigmoid} activation function on the hidden layer units.
All the models are trained using mini-batches of size five. 
The maximum number of epochs is set to 50, but we also use an early stopping
criterion on the model's accuracy on the validation split (i.e.\ training is
interrupted if the validation accuracy doesn't improve over five consecutive
epochs).
We implement all the models in \emph{Keras} with \emph{TensforFlow} as 
backend. 
All the TL and MTL models are trained with the same hyperparameters of the STL
model.\footnote{
\href{https://github.com/ltgoslo/fun-nom}{github.com/ltgoslo/fun-nom}}

\subsection{Transfer Learning Models}
\label{ssec:tl}

Transfer learning in our experiments amounts to training an STL model on \pcedt
relations, for example, and then using (some of) its weights to initialize
another model for \nombank relations.
Given the architecture of the neural classifier described in \sref{sec:stl},
we identify three ways to implement TL:
1) \textbf{TL\textsubscript{E}}: Transfer of the embedding layer weights, 
2) \textbf{TL\textsubscript{H}}: Transfer of the hidden layer weights, and
3) \textbf{TL\textsubscript{EH}}: Transfer of both the embedding and hidden
layer weights. 
Furthermore, we distinguish between transfer learning from \pcedt to \nombank
and vice versa; that is, either task can be used as main task or auxiliary task.
Hence, we either start by training on \nombank and use the weights
of the corresponding transfer layer to initialize the \pcedt model or the other
way around.
In total, this leads to six setups, as shown in \tref{tab:transfer}.
Note that we do not apply TL (or MTL) on the output layer because it is task- or
dataset-specific \cite{Mou:Men:Yan:16}.

\subsection{Multi-Task Learning Models}
\label{ssec:mtl}

In MTL, we train one model simultaneously to learn both \pcedt and \nombank
relations, and therefore all the MTL models have two objective functions and two
output layers.
We implement two MTL setups: \textbf{MTL\textsubscript{E}}, which consists of
a shared embedding layer but two task-specific hidden layers, and 
\textbf{MTL\textsubscript{F}}, 
which, apart from the output layer, does not have task-specific layers, i.e.\
both the embedding and hidden layers are shared.
We distinguish between the auxiliary and main tasks 
based on which validation accuracy (\nombank's\ or \pcedt's) is monitored by the
early
stopping criterion.
Hence we end up with a total of four MTL models as shown in 
\tref{tab:mtl}.

\section{Experimental Results}
\label{sec:results}

Tables \ref{tab:transfer} and \ref{tab:mtl} present the accuracies of the
different TL and MTL models on the development and test splits in \nombank
and \pcedt. 
The top row in both tables shows the accuracy of the STL model. 
All the models were trained on the training split only.
There are several observations one can draw from these tables.
First, the accuracy of the STL models drops when the models are evaluated on the
test split, whether on \nombank or \pcedt.
Second, all the TL models achieve better accuracy
on the test split of \nombank even though transfer learning does not remarkably
improve accuracy on the development split of the same dataset.
The MTL models, especially MTL\textsubscript{F}, have a negative effect on the
development accuracy of \nombank, but we still see the same improvement, as in
TL, on the test split.
Third, both the TL and MTL models exhibit less consistent effects on
\pcedt (on both the development and test splits) compared to
\nombank; for example, all the TL models lead to about 1.25 points absolute
improvement in accuracy on \nombank, whereas in \pcedt TL\textsubscript{E} is clearly better
than the other two TL models (TL\textsubscript{E} improves over the STL accuracy
by 1.37 points).

\begin{table}[t!]
\small
\begin{center}
\begin{tabular}{@{}l | c | c || c | c@{}}
Model & \multicolumn{2}{c||}{\nombank} & \multicolumn{2}{c}{\pcedt} \\ 
     & Dev & Test & Dev & Test \\ \hline
STL  & 78.15 & 76.75 & 58.80 & 56.05 \\
TL\textsubscript{E} & 78.37 & \bf{78.05} & 59.57 & \bf{57.42} \\
TL\textsubscript{H} & 78.15 & 78.00 & 59.24 & 56.51 \\
TL\textsubscript{EH} & \bf{78.48} & 78.00 & \bf{59.89} & 56.68 \\
\hline
\end{tabular}
\end{center}
\caption{\label{tab:transfer} Accuracy (\%) of the transfer learning models.}
\end{table}

\begin{table}[t!]
\small
\begin{center}
\begin{tabular}{@{}l | c | c || c | c@{}}
Model & \multicolumn{2}{c||}{\nombank} & \multicolumn{2}{c}{\pcedt} \\ 
     & Dev & Test & Dev & Test \\ \hline
STL  & \bf{78.15} & 76.75 & 58.80 & 56.05 \\
MTL\textsubscript{E} & 77.93 & 78.45 & \bf{59.89} & \bf{56.96}  \\
MTL\textsubscript{F} & 76.74 & \bf{78.51} & 58.91 & 56.00 \\
\hline
\end{tabular}
\end{center}
\caption{\label{tab:mtl} Accuracy (\%) of the MTL models.}
\end{table}

Overall, the accuracy of the STL models drops when evaluated on the test split
of \nombank and \pcedt (in comparison to their accuracy on the development
split); this might be an indicator of overfitting, especially because we select
the model that performs best on the development split in our stopping criterion.
Both TL and MTL, on the other hand, improve accuracy on the test splits, even
though the same stopping criterion was used for STL, TL and MLT. 
We interpret this result as improvement in the models' generalization ability.
However, given that these improvements are relatively small, we next take a
closer look at the results to understand if and how TL and MTL help.

\section{Results Analysis}
\label{ssec:analysis}

This section presents a systematic analysis of the performance of the models
based on insights from the dataset used in the experiments as well as the
classification errors of the models.
The discussion in the following sections is based on the results on the test
split rather than the development split, primarily because the former is larger
in size.\footnote{One can also argue that result analysis on the test split
is stricter than on the validation split. 
While using an early stopping criterion based on the validation data can help
prevent overfitting on the training data, we still choose a model that achieves
the best accuracy on the validation split. 
In addition, it's unclear if early stopping helps when the validation split is 
not fully representative of the problem \cite{Prechelt:12}.} 

\subsection{Relation Distribution}
\label{ssec:analysis-dataset}

\begin{figure*} 
\centering 
\begin{minipage}{.5\textwidth}
  \includegraphics[width=1.\linewidth]{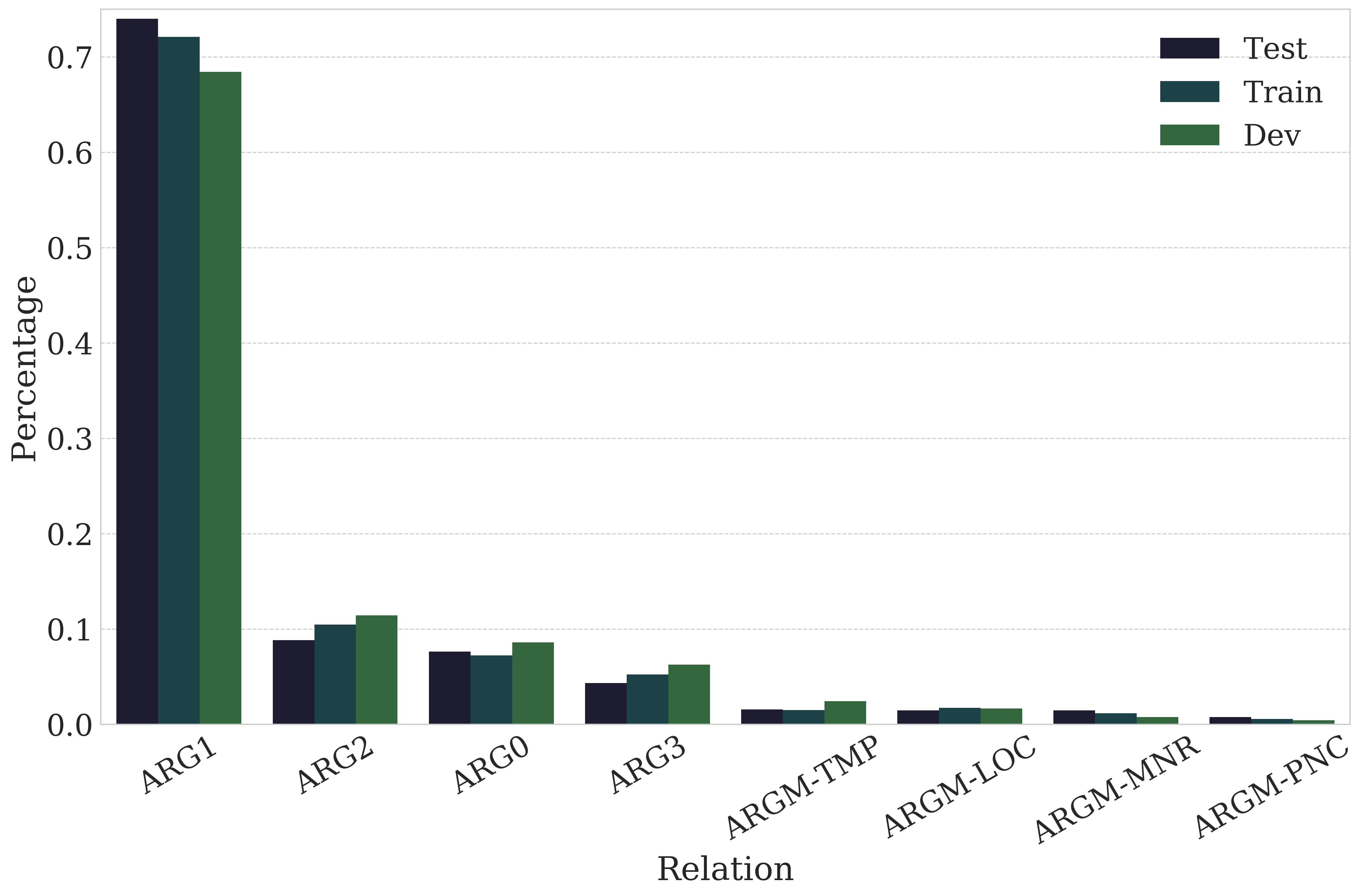}
\end{minipage}%
\begin{minipage}{.5\textwidth}   
  \centering
  \includegraphics[width=1.\linewidth]{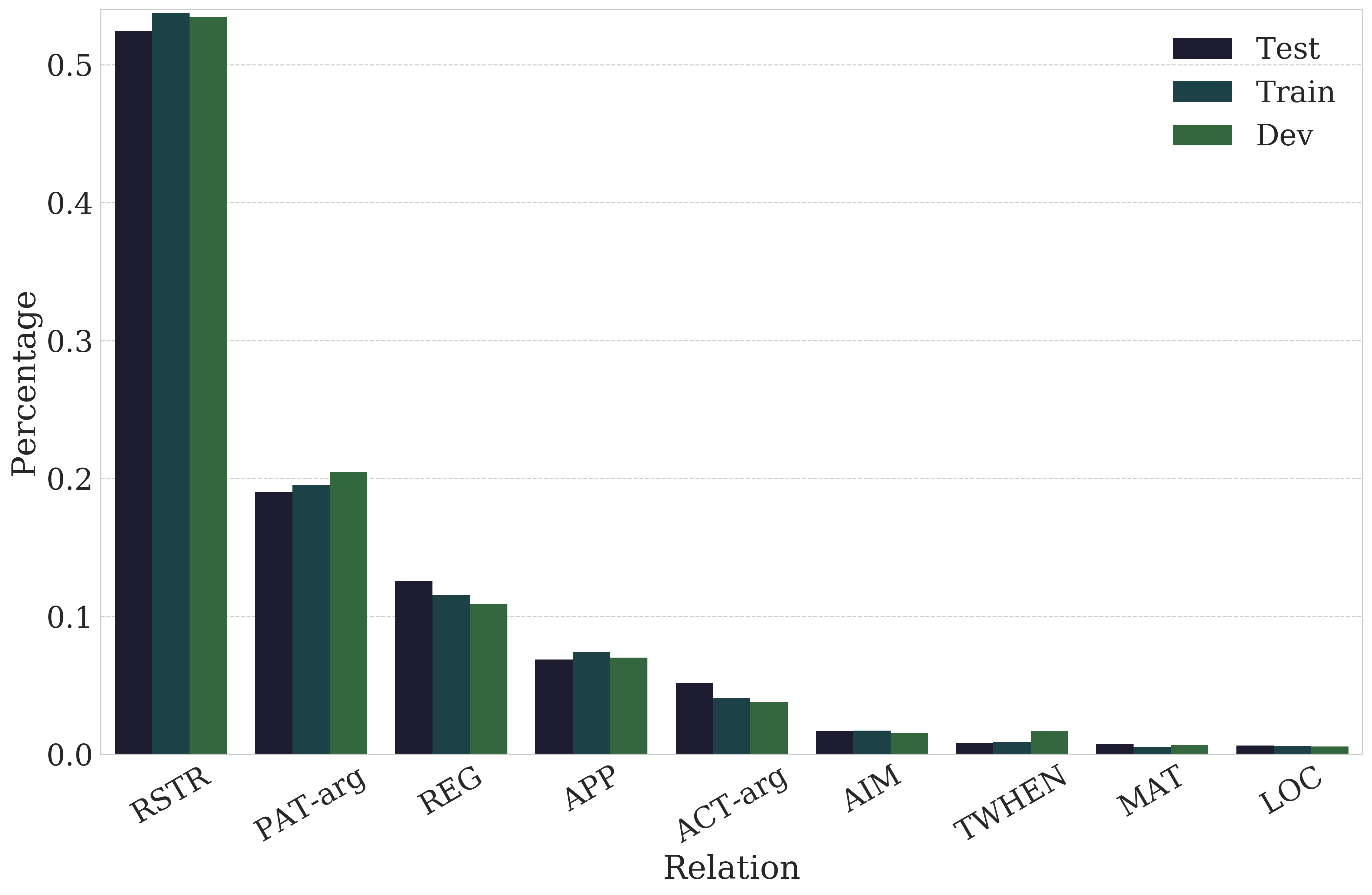}
  \end{minipage}
    \caption{Distribution of \nombank relations (left) and \pcedt relations 
    (right)}
  \label{fig:relation-distribution} 
\end{figure*}

To demonstrate the difficulty of the problem at hand, we plot the
distribution of the most frequent relations in \nombank and \pcedt across the
three data splits in \fref{fig:relation-distribution}.
We find that almost 71.18\% of the relations in the \nombank training split are
of type \texttt{ARG1} (proto-typical patient), and 52.20\% of
the \pcedt relations are of type \texttt{RSTR} (an underspecified adnominal
modifier).
Such highly skewed distributions of the relations makes learning some of the other
relations more difficult, if not impossible in some cases. 
In fact, of the 15 \nombank relations observed in the test split, five relations
are never predicted by any of the STL, TL and MTL models, and of the 26
\pcedt relations observed in the test split only six are predicted.
That said, the non-predicted relations are extremely infrequent
in the training split (e.g.\ 23 \pcedt functors occur less than 20 times in the
training split), and it is therefore questionable if an ML model will be able to
learn them under any circumstances.

From this imbalanced distribution of relations, it immediately follows that
accuracy alone, as evaluation measure, is not sufficient to identify the best
performing model.
Therefore, in the following section we report, and analyze, the \fone scores of
the predicted \nombank and \pcedt relations across all the STL, TL and MTL
models.

\subsection{Per-Relation \fone Scores}
\label{ssec:per-label}

Tables \ref{tab:nom-per-class} and \ref{tab:pcedt-per-class} show the
per-relation \fone scores for \nombank and \pcedt, respectively.
Note that we only include the results for the relations that are actually
predicted by at least one of the models.

\setlength\tabcolsep{3pt}
\begin{table}[t!]
\small
\begin{center} 
\begin{tabular}{@{}l | c  c  c  c  c  c  c@{}}
     & A0 & A1 & A2 & A3 & LOC & MNR & TMP \\ 
     \emph{Count} & 132 & 1282 & 153 & 75 & 25 & 25 & 27\\ \hline
STL   & 49.82 & 87.54 & \bf{45.78} & 60.81 & 28.57 & 29.41 & \bf{66.67}\\ 
TL\textsubscript{E}  & \bf{55.02} & 87.98 & 41.61 & 60.14 & 27.91 & 33.33 & 63.83 \\ 
TL\textsubscript{H}  & 54.81 & 87.93 & 42.51 & 60.00 & 25.00 & \bf{35.29} & 65.31 \\ 
TL\textsubscript{EH} & 53.62 & 87.95 & 42.70 & 61.11 & 29.27 & 33.33 & 65.22 \\ 
MTL\textsubscript{E} & 54.07 & 88.34 & 42.86 & 61.97 & \bf{30.00} & 28.57 & \bf{66.67} \\ 
MTL\textsubscript{F} & 53.09 & \bf{88.41} & 38.14 & \bf{62.69} & 00.00 & 00.00 & 52.17 \\ 
\hline

\end{tabular}
\end{center}
\caption{\label{tab:nom-per-class} Per-label \fone score on the \nombank test
split.}
\end{table}

\setlength\tabcolsep{3pt}
\begin{table}[t!]
\small
\begin{center}
\begin{tabular}{@{}l | c  c  c  c  c  c@{}}
     & ACT & TWHEN & APP & PAT & REG & RSTR\\ 
\emph{Count} & 89 & 14 & 118 & 326 & 216 & 900\\ \hline
STL   & 43.90 & 42.11 & 22.78 & 42.83 & 20.51 & 68.81 \\
TL\textsubscript{E}   & 49.37 & \bf{70.97} & 27.67 & 41.60 & \bf{30.77} & \bf{69.67} \\
TL\textsubscript{H}   & 53.99 & 62.07 & 25.00 & \bf{43.01} & 26.09 & 68.99 \\
TL\textsubscript{EH}  & 49.08 & 64.52 & \bf{28.57} & 42.91 & 28.57 & 69.08 \\
MTL\textsubscript{E} & \bf{54.09} & 66.67 & 24.05 & 42.03 & 27.21 & 69.31 \\
MTL\textsubscript{F} & 47.80 & 42.11 & 25.64 & 40.73 & 19.22 & 68.89 \\
\hline
\end{tabular}
\end{center}
\caption{\label{tab:pcedt-per-class} Per-label \fone score on the \pcedt test 
split.}
\end{table}

We observe several interesting patterns in Tables \ref{tab:nom-per-class} and 
\ref{tab:pcedt-per-class}.
First, the MTL\textsubscript{F} model seems to be confusing for both datasets: it leads to
substantially degraded \fone scores on four \nombank relations, including 
the locative modifier \texttt{ARGM-LOC} and manner modifier \texttt{ARGM-MNR} 
(shortened to \texttt{LOC} and \texttt{MNR} in \tref{tab:nom-per-class}) which
the model is no longer able to predict.
The same model has the worst \fone score, compared to all other models, for two
\pcedt relations, \texttt{REG} (which expresses a circumstance) and 
\texttt{PAT} (patient).
Given that the MTL\textsubscript{F} model achieves the highest accuracy on the
\nombank test split (cf.\ \tref{tab:mtl}), it becomes all the more evident
that mere accuracy scores are not enough to judge the utility of TL and MTL for
this task (and dataset).

Second, with the exception of the MTL\textsubscript{F} model, all the TL and MTL models
consistently improve the \fone score of all the \pcedt relations except 
\texttt{PAT}.
Most notably, the \fone scores of the relations \texttt{TWHEN} and \texttt{ACT}
see a remarkable boost, compared to other \pcedt relations, when
only the embedding layer's weights are shared (MTL\textsubscript{E}) or
transfered (TL\textsubscript{E}).
This result can be partly explained by looking at the correspondence matrices
between \nombank arguments and \pcedt functors shown in Tables
\ref{tab:correspond-nom-pcedt} and \ref{tab:correspond-pcedt-nom}, which
show how the \pcedt functors map to \nombank arguments in the
training split (\tref{tab:correspond-pcedt-nom}) and the other way around
(\tref{tab:correspond-nom-pcedt}).
From \tref{tab:correspond-pcedt-nom}, we see that 80\% of the compounds annotated
as \texttt{TWHEN} in \pcedt were annotated as \texttt{ARGM-TMP} in \nombank.
In addition, 47\% of \texttt{ACT} (Actor) relations map to \texttt{ARG0} 
(Proto-Agent) in \nombank,
even though this mapping is not as clear as one would have hoped, it is still
relatively high if we consider how other \pcedt relations map to \texttt{ARG0}.
The correspondence matrices also show that the assumed theoretical
similarities between the \nombank and \pcedt relations do not always hold.
Nonetheless, even such `imperfect' correspondence can provide a `training
signal' that help the TL and MTL models learn relations such as \texttt{TWHEN}
and \texttt{ACT}. 

Since the TL\textsubscript{E} model outperforms STL on predicting 
\texttt{REG} by ten absolute points, we inspected all the \texttt{REG} compounds that were
correctly classified by the TL\textsubscript{E} model but were misclassified by the STL model
and found that the latter misclassified them as \texttt{RSTR} which indicates
that TL from \nombank helps the TL\textsubscript{E} model recover from the STL's
over-generalization in
\texttt{RSTR} prediction.

The two \nombank relations that receive the highest boost in \fone score (about
five absolute points) are \texttt{ARG0} and \texttt{ARGM-MNR}, but the
improvement in the latter relation corresponds to only one more compound
which might well be predicted correctly by chance.
Overall, TL and MTL from \nombank to \pcedt is more helpful than the other way
around.
One way to explain this is considering the first rows in Tables
\ref{tab:correspond-pcedt-nom} and \ref{tab:correspond-nom-pcedt}, where we
see that five \pcedt relations (including the four most frequent ones) map
to \texttt{ARG1} in \nombank in more than 60\% of the cases for each relation.
This means that the weights learned to predict \pcedt relations offer little or
no inductive bias for \nombank relations.
Whereas if we consider the mapping from \nombank to \pcedt, we see that even
though many \nombank arguments map to \texttt{RSTR} in \pcedt the percentages
are lower, and hence the mapping is more `diverse' (i.e.\ discriminative) which
seems to help the
TL and MTL models learn the less frequent \pcedt relations.

\begin{table}[t!]
\small
\begin{center}
\begin{tabular}{@{}l | c c c c c c c c@{}}
& \texttt{RSTR} & \texttt{PAT} & \texttt{REG} & \texttt{APP} & \texttt{ACT} & \texttt{AIM} & \texttt{TWHEN} \\
\hline
\texttt{A1} & 0.70 & 0.90 & 0.78 & 0.62 & 0.47 & 0.65 & 0.10 \\
\texttt{A2} & 0.11 & 0.05 & 0.10 & 0.21 & 0.03 & 0.12 & 0.03 \\
\texttt{A0} & 0.06 & 0.01 & 0.04 & 0.13 & 0.47 & 0.07 & -\\
\texttt{A3} & 0.06 & 0.02 & 0.06 & 0.02 & 0.01 & 0.06 & - \\
\texttt{LOC} & 0.02 & 0.01 & 0.00 & 0.01 & 0.01 & 0.01 & 0.02  \\
\texttt{TMP} & 0.01 & - & 0.00 & 0.00 & - & - & 0.80  \\
\texttt{MNR} & 0.02 & 0.00 & 0.00 & - & 0.01 & - & - \\
\hline
\emph{Count} & 3617 &  1312 &   777 &   499 &273 &   116 & 59 \\
\end{tabular}
\end{center}
\caption{\label{tab:correspond-pcedt-nom} Correspondence matrix between \pcedt
functors and \nombank arguments. Slots with `-' mean zero, 0.00 is a very small
number but not zero.}
\end{table}

\begin{table}[t!]
\small
\begin{center}
\begin{tabular}{@{}l | c c c c c c c@{}}

& \texttt{A1} & \texttt{A2} & \texttt{A0} & \texttt{A3} & \texttt{LOC} & \texttt{TMP} & \texttt{MNR} \\
\hline
\texttt{RSTR} & 0.51  & 0.54  & 0.47  & 0.63  & 0.66  & 0.36  & 0.78 \\
\texttt{PAT} & 0.24  & 0.09  & 0.03  & 0.08  & 0.07  & -  & 0.05 \\
\texttt{REG} & 0.12  & 0.11  & 0.07  & 0.13  & 0.02  & 0.01  & 0.01 \\
\texttt{APP} & 0.06  & 0.14  & 0.13  & 0.03  & 0.05  & 0.01  & -\\
\texttt{ACT} & 0.03  & 0.01  & 0.26  & 0.01  & 0.03  & - & 0.03 \\
\texttt{AIM} & 0.02  & 0.02  & 0.02  & 0.02  & 0.01  & - & - \\
\texttt{TWHEN} & 0.00  & 0.00  & -  & -    & 0.01  & 0.46  & - \\
\hline
\emph{Count} &  4932 &   715 &   495 &   358 &   119 &   103 & 79 

\end{tabular}
\end{center}
\caption{\label{tab:correspond-nom-pcedt} Correspondence matrix between \nombank
arguments and \pcedt functors.}
\end{table}

For completeness, we investigate why the \pcedt functor \texttt{AIM} is never
predicted even though it is more frequent than \texttt{TWHEN} (cf.\ 
\fref{fig:relation-distribution}).
We find that \texttt{AIM} is almost always misclassifed as \texttt{RSTR} 
by all the models.
Furthermore, we discover that \texttt{AIM} and \texttt{RSTR} have the highest
lexical overlap in the training set among all other pairs of relations in \pcedt:
78.35\% of the left constituents and 73.26\% of the right constituents of the
compounds annotated as
\texttt{AIM} occur in other compounds annotated as \texttt{RSTR}.
This explains why none of the models manage to learn the relation \texttt{AIM}
but raises a question about the models' ability to learn relational
representations; we pursue this question further in \sref{ssec:generalization}.

Finally, to clearly demonstrate the benefits of TL and MTL for \nombank and
\pcedt, we report the \fone macro-average scores in \tref{tab:macro-avg} (which
is arguably the appropriate evaluation measure for imbalanced
classification problems).
Note that the relations that are not predicted by any of the models are not
included in computing the macro-average.
From \tref{tab:macro-avg} it becomes crystal clear that TL and MTL on the
embedding layer yield remarkable improvements for \pcedt with about 7--8
absolute points increase in macro-average \fone, in contrast to just
$0.65$ in the best case on \nombank.

\begin{table}[t!]
\small
\begin{center}
\begin{tabular}{@{}l | c | c@{}}
Model & \nombank & \pcedt \\ \hline
STL   & 52.66 & 40.15 \\
TL\textsubscript{E}  & 52.83 & \bf{48.34} \\
TL\textsubscript{H}  & 52.98 & 46.52 \\
TL\textsubscript{EH} & \bf{53.31} & 47.12 \\
MTL\textsubscript{E} & 53.21 & 47.23 \\
MTL\textsubscript{F} & 42.07 & 40.73 \\
\hline
\end{tabular}
\end{center}
\caption{\label{tab:macro-avg} Macro-average \fone score on the test split.}
\end{table}

\subsection{Generalization on Unseen Compounds}
\label{ssec:generalization}

Now we turn to analyze the models' ability to generalize over compounds unseen 
in the training split.
Recent work by \newcite{Dima:16} and \newcite{Ver:Wat:18} suggest that the gains
achieved in noun--noun compound interpretation using word embeddings
and somewhat similar neural classification models are in fact a by-product of
\emph{lexical memorization} \cite{Lev:Rem:Bie:15}; 
in other words, the classification models learn that a specific set of
nouns is a strong indicator of a specific relation. 
Therefore, in order to gauge the role of lexical memorization in our
models also, we quantify the number of unseen compounds that the STL, TL
and MTL models predict correctly.

We distinguish between `partly' and `completely' unseen compounds.
A compound is considered `partly' unseen if one of its constituents (right or
left) is not seen in the training data at all.
A completely unseen compound is one whose left \emph{and}
right constituent are not seen in the training data (i.e.\ completely unseen
compounds are the subset of compounds in the test split that have zero lexical
overlap with the training split).
Overall, almost $20\%$ of the compounds in the test split have an unseen
left constituent, about $16\%$ of the compounds have unseen right constituent
and $4\%$ are completely unseen.
In \tref{tab:generalization}, we compare the performance of the different models
on these three groups in terms of the proportion of compounds a model 
\emph{misclassifies} in each group. 

\setlength\tabcolsep{5pt}
\begin{table}[t!]
\small
\begin{center}
\begin{tabular}{@{}l | c  c  c | c  c  c@{}}
Model & \multicolumn{3}{c|}{\nombank} & \multicolumn{3}{c}{\pcedt} \\ 
     & L & R & L\&R & L & R & L\&R\\ 
\emph{Count} & 351 & 286 & 72 & 351 & 286 & 72\\  \hline
STL  & 27.92 & 39.51 & 50.00 & 45.01 & 47.55 & 41.67 \\
TL\textsubscript{E}  & 25.93 & 36.71 & 48.61 & \bf{43.87} & 47.55 & 41.67 \\
TL\textsubscript{H}  & 26.21 & 38.11 & 50.00 & 46.15 & 49.30 & 47.22 \\
TL\textsubscript{EH} & 26.50 & 38.81 & 52.78 & 45.87 & 47.55 & 43.06 \\
MTL\textsubscript{E} & 24.50 & \bf{33.22} & \bf{38.89} & 44.44 & \bf{47.20} & 43.06 \\
MTL\textsubscript{F} & \bf{22.79} & 34.27 & 40.28 & 44.16 & 47.90 & \bf{38.89} \\
\hline
\end{tabular}
\end{center}
\caption{\label{tab:generalization} Generalization error on the subset of
unseen compounds in the test split. L: Left constituent. R: Right constituent.
L\&R: Completely unseen.}
\end{table}

\begin{figure*} 
\centering 
\begin{minipage}{.5\textwidth}
  \includegraphics[width=1.\linewidth]{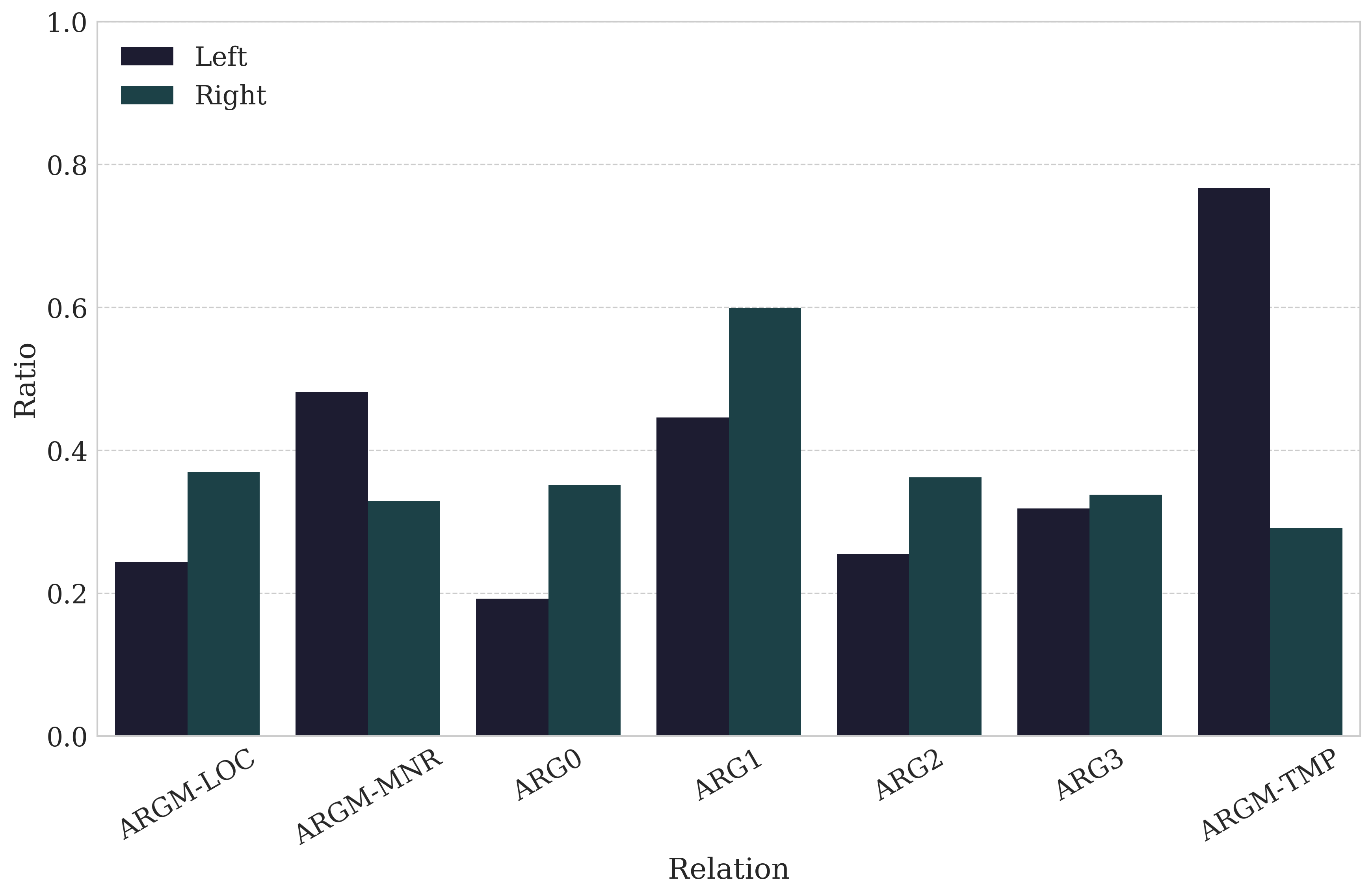}
\end{minipage}%
\begin{minipage}{.5\textwidth}   
  \centering
  \includegraphics[width=1.\linewidth]{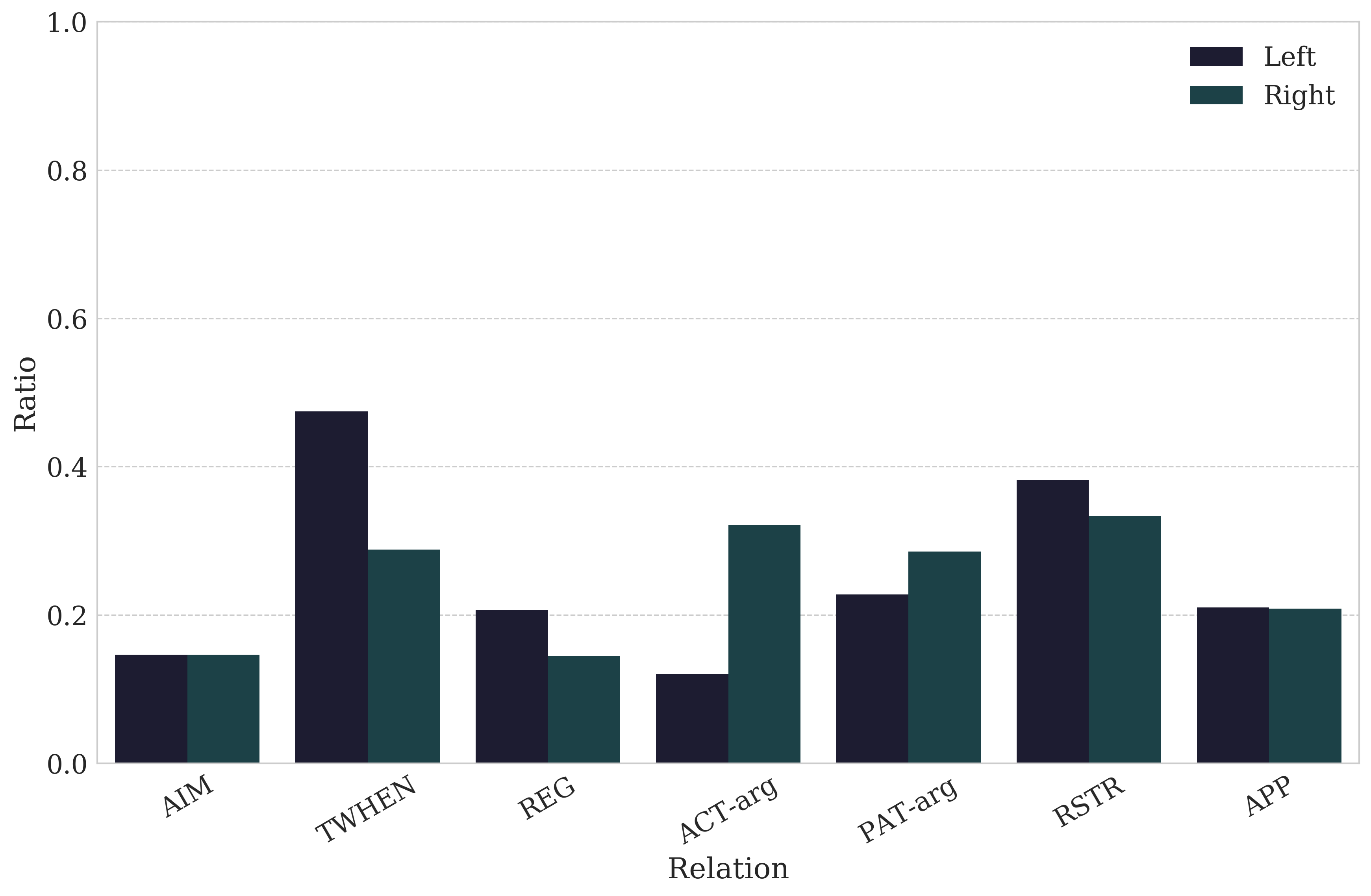}
  \end{minipage}
    \caption{Ratio of relation-specific constituents in \nombank
    (left) and \pcedt (right).}
  \label{fig:const-per-rel} 
\end{figure*}

From \tref{tab:generalization} we see that TL and MTL reduce the \nombank
generalization error in all cases, except TL\textsubscript{H} and 
TL\textsubscript{EH} on completely unseen compounds; the latter leads to
higher generalization error.
The MTL models lead to the biggest error reduction across the three types of
unseen compounds; MTL\textsubscript{E} leads to about six points error reduction
on compounds with unseen right constituent and eleven points on completely
unseen ones, and MTL\textsubscript{F} reduces the error on 
unseen left constituent by five points. 
Note, however, that these results have to be read together with the
\emph{Count} row in \tref{tab:generalization} to get a
complete picture.
For instance, an eleven-point decrease in error on completely unseen
compounds amounts to eight compounds.
In \pcedt, the largest error reduction on unseen left constituents is 1.14
points which amounts to four compounds, 0.35 (just one compound) on unseen
right constituents and 2.7 (or two compounds) on completely unseen
compounds.

Since we see larger reductions in the generalization error in \nombank, we
manually inspect the compounds that led to these reductions; i.e.\
we inspect the distribution of relations in the set of the correctly predicted
unseen compounds. 
The MTL\textsubscript{E} model reduces the generalization error on completely
unseen compounds by a total of eight compounds compared to the STL model, but
seven of these compounds are annotated with \texttt{ARG1} which is the most
frequent relation in \nombank.
When it comes to the unseen right constituents, the 24 compounds 
MTL\textsubscript{E} improves on consist of 18 \texttt{ARG1} compounds, 5 
\texttt{ARG0} compounds and one \texttt{ARG2} compound. 
We see a similar pattern upon inspecting the gains of the TL\textsubscript{E} model; 
where most of the improvement arises from predicting more \texttt{ARG1} and \texttt{ARG0}
correctly.

The majority of the partly or completely unseen compounds that were
misclassified by all models are \emph{not} of type \texttt{ARG1} in \nombank
or \texttt{RSTR} in \pcedt.
This, together with the fact that the correctly predicted unseen compounds
are annotated with the most frequent relations, indicate that the classification
models rely on lexical memorization to learn the interpretation of
compound relations. 

Finally, to complement our understanding of the effect of lexical memorization,
we plot the ratio of relation-specific constituents in \nombank and \pcedt in 
\fref{fig:const-per-rel}.
We define relation-specific constituents as left or right constituents that
only occur with one specific relation in the training split, and their ratio
is simply their proportion in the overall set of left or right constituents per
relation. 
Looking at \fref{fig:const-per-rel}, we see that 
\nombank relations have higher ratios of relation-specific constituents in
comparison to \pcedt, which arguably makes learning the former comparatively
easier if the model is only to rely on lexical memorization.
Furthermore, \texttt{ARGM-TMP} in \nombank and \texttt{TWHEN} in \pcedt
stand out from other relations in \fref{fig:const-per-rel}, which
are also the two relations with the second highest \fone score in their
respective dataset---except in STL on \pcedt (cf.\ Tables 
\ref{tab:nom-per-class} and \ref{tab:pcedt-per-class}).
Lexical memorization is, therefore, the most likely explanation of such
relatively high \fone scores.
We also observe some correlation between lower ratios of relation-specific
constituents and relatively low \fone scores, e.g.\ \texttt{APP} and 
\texttt{REG} in \pcedt. 
Based on these observations, we cannot rule out that our models exhibit some
degree of lexical memorization effects, even though manual result analysis also
reveals `counter-examples' where the models generalize and make correct
predictions where lexical memorization is impossible. 

\section{Conclusion}
\label{sec:conc}

Transfer and multi-task learning for NLP currently receive a lot of
attention, but for the time being there remains considerable
uncertainty about which task properties and experimental settings
actually are effective.
In this work, we seek to shed light on the utility of TL and MTL
perspectives on the semantic interpretation of noun--noun compounds.
Through a comprehensive series of minimally contrasting experiments and
in-depth analysis of results and prediction errors, we demonstrate the
ability of both TL and MTL to mitigate the challenges of class
imbalance and substantially improve prediction of low-frequency
relations.
In a nutshell, our TL and in particular MTL models make quantitatively
and qualitatively better predictions, especially so on the `hardest'
inputs involving at least one constituent not seen in the training
data---but clear indicators of remaining `lexical memorization' effects
arise from our error analysis of unseen compounds.

In general, transfer of representations or sharing across tasks is most
effective at the embedding layers, i.e.\ the model-internal
representation of the two compound constituents involved.
In multi-task learning, full sharing of the model architecture across tasks
worsens the model's ability to generalize on the less frequent relations.

We experience the dataset by \newcite{Fares:2016} as an interesting opportunity
for innovative neural approaches to compound interpretation, as it
relates this sub-problem to broad-coverage semantic role labeling or semantic
dependency parsing in \pcedt and \nombank.
In future work, we plan to incorporate other NLP tasks defined over these
frameworks to learn noun--noun compound interpretation using TL and MTL.
Such tasks include semantic role labeling of nominal predicates in
\nombank annotations as well as verbal predicates in \texttt{PropBank} 
\cite{Kin:Pal:02}.


\bibliographystyle{acl_natbib_nourl}
\bibliography{local,ltg}

\end{document}